\pdfoutput=1

\documentclass[11pt]{article}

\usepackage{emnlp2021}

\usepackage{times}
\usepackage{latexsym}

\usepackage[T1]{fontenc}

\usepackage[utf8]{inputenc}

\usepackage{microtype}

\usepackage{tikz-dependency}
\usepackage{tikz}
\usepackage{tikz-qtree}
\usepackage{latexsym}
\usepackage{amssymb}
\usepackage{amsmath}
\usepackage{arydshln}
\newcommand{\added}[1]{\textcolor{black}{#1}}

\title{Reducing Discontinuous to Continuous Parsing with Pointer Network Reordering}

\author{Daniel Fern\'{a}ndez-Gonz\'{a}lez \and Carlos G\'{o}mez-Rodr\'{i}guez\\
	Universidade da Coru\~{n}a, CITIC\\
	FASTPARSE Lab, LyS Group \\
Depto. de Ciencias de la Computaci\'{o}n y Tecnolog\'{i}as de la Informaci\'{o}n \\
	Campus de Elvi\~{n}a, s/n, 15071 A Coru\~{n}a, Spain \\
  {\tt d.fgonzalez@udc.es}, {\tt carlos.gomez@udc.es}\\}

\date{}

\begin{document}
\maketitle
\begin{abstract}
Discontinuous constituent parsers have always lagged behind continuous approaches in terms of accuracy and speed,
as
the presence of constituents 
with discontinuous yield
introduces extra complexity to the 
task.
However, 
a discontinuous 
tree can be converted into a continuous variant by 
reordering tokens. Based on that, we propose to reduce discontinuous 
parsing to a continuous problem, which can then be directly solved by any off-the-shelf continuous parser. To that end, we develop a Pointer Network
capable of accurately generating 
the continuous 
token arrangement for a given input sentence and define a bijective function to recover the original order. Experiments 
on the main benchmarks with 
two 
continuous parsers prove that our approach 
is on par in accuracy with purely discontinuous state-of-the-art algorithms, 
but considerably faster.

\end{abstract}

\section{Introduction}
\begin{figure*}[t]
\includegraphics[width=\textwidth]{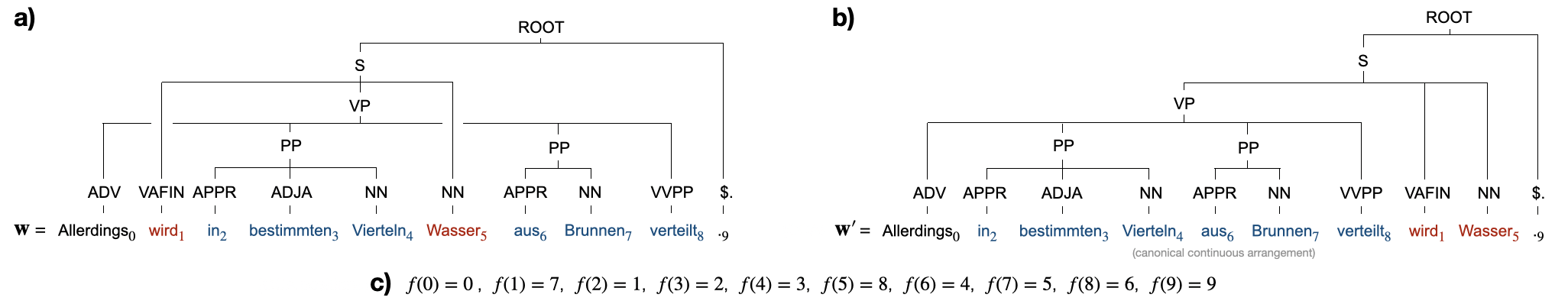}
\caption{\textbf{a)} Discontinuous constituent tree for a sentence in German NEGRA development split, \textbf{b)} its canonical continuous arrangement and \textbf{c)} conversion of original positions into absolute CCA positions through function $f$ implemented by a Pointer Network.}
\label{fig:trees}
\end{figure*}

Discontinuous phrase-structure trees (with crossing branches like the one in Figure~\ref{fig:trees}(a)) are crucial for fully representing the wide range of syntactic phenomena present in human languages such as
long-distance extractions, dislocations or cross-serial dependencies, among others.

Although continuous approaches ignore these linguistic phenomena by, for instance, removing them from the original treebank (a common practice in the 
Penn Treebank \cite{marcus93}), there exist different algorithms that can handle discontinuous parsing. Currently, 
we can highlight (1) those based in Linear Context-Free Rewriting Systems (LCFRS) \cite{VijWeiJos87}, which allow exact CKY-style parsing of discontinuous structures at a high computational cost
\cite{gebhardt-2020-advances,morbitz2020supertaggingbasedNAACL}; (2) a variant of 
the former
that, while still making use of LCFRS formalisms,  
increases parsing speed 
by implementing a span-based scoring algorithm \cite{stern-etal-2017-minimal} and not explicitly defining a set of rules \cite{stanojevic-steedman-2020-span, Corro2020SpanbasedDC}; (3) 
transition-based parsers that 
deal
with discontinuities by adding a specific transition in charge of changing token order \cite{versley2014,maier-2015-discontinuous,maier2016,stanojevic-alhama-2017-neural,coavoux2017}
or by designing new data structures that allow interactions between already-created non-adjacent subtrees \cite{coavoux2019a,coavoux2019b}; and, finally, (4)  several approaches that reduce discontinuous constituent parsing to a simpler problem, 
converting it, for instance, into a non-projective dependency parsing task \cite{fernandez-gonzalez-martins-2015-parsing,DiscoPointer} or into a sequence labelling problem \cite{vilares-gomez-rodriguez-2020-discontinuous}. \added{In (4),
we can also include the solutions proposed by \citet{boyd-2007-discontinuity} and \citet{Versley2016}, which transform discontinuous treebanks into continuous variants where discontinuous constituents are encoded by creating additional constituent nodes and extending the original non-terminal label set (following a pseudo-projective technique \cite{nivre-nilsson-2005-pseudo}), to then be processed by continuous 
parsing models and discontinuities recovered in a post-processing step.} 

It is well known that discontinuities are inherently related to the order of tokens in the sentence, and a discontinuous tree can be transformed into a continuous one by \added{just} reordering the words \added{and without including additional structures}, an idea that has been exploited in \added{practically all} transition-based parsers 
and other approaches  \cite{vilares-gomez-rodriguez-2020-discontinuous}. However, in these models the reordering process is tightly integrated and inseparable from the parsing process. 

\added{Likely due to the lack of accurate models to accomplish 
reordering in isolation, we are not aware of any approach framed in (4) that explicitly reduces discontinuous constituent parsing into a continuous problem, keeping the original set of constituent nodes and solving it with a completely independent
continuous parser that does not have to deal with an extended label set. Please note that existing approaches that perform 
discontinuous-to-continuous conversion, such as \cite{boyd-2007-discontinuity} and \cite{Versley2016}, not only modify the original discontinuous tree by including artificial constituent nodes and enlarging its label scheme 
(probably penalizing 
parsing performance), but they are not able to fully recover the original discontinuous tree due to limitations of the proposed 
encodings.
}

In this paper, we study the (\added{fully} reversible) discontinuous-to-continuous conversion \added{by token reordering} and how any off-the-shelf continuous parser can be directly applied without any further adaptation \added{or extended label set}. To undertake the \added{independent} token reordering, we rely on a 
Pointer Network architecture \cite{Vinyals15} that can accurately relocate those tokens causing discontinuities in the sentence to new positions, generating 
new sentences that can be directly parsed by any continuous parser. We test our approach
\footnote{Source code available at \url{https://github.com/danifg/Pointer-Network-Reordering}.}
with two 
continuous algorithms \cite{kitaev-etal-2019-multilingual,attachjuxtapose}
on three widely-used discontinuous treebanks, obtaining remarkable accuracies and outperforming 
current state-of-the-art discontinuous parsers 
in 
terms of 
speed.

\section{Pointer Network Reordering}
\subsection{Continuous Canonical Arrangement}

Let $\mathbf{w} = w_0, \dots ,w_{n-1}$ be an input sentence of $n$ tokens, and $t$ a discontinuous constituent tree for $\mathbf{w}$. We are interested in a permutation (reordering) $\mathbf{w'}$ of $\mathbf{w}$ that turns $t$ into a continuous tree $t'$. While there can be various permutations that achieve this for a given tree, we will call \textit{continuous canonical arrangement (CCA)} of $\mathbf{w}$ and $t$ the permutation obtained by placing the tokens of $\mathbf{w}$ in the order given by an in-order traversal of $t$.

This permutation defines a bijective function, $f: \{0, \ldots, n-1\} \rightarrow \{0, \ldots, n-1\}$, mapping each token at position $i$ in $\mathbf{w}$ to its new \textit{CCA position} $j$ in $\mathbf{w'}$. Then, $\mathbf{w'}$ can be parsed by a continuous parser and, by keeping track of $f$ (i.e., storing original token positions), it is trivial to recover the discontinuous tree by applying its inverse $f^{-1}$.
The challenge is in accurately predicting the CCA positions for a given sentence $\mathbf{w}$ (i.e. learning $f$) \emph{without} knowing the parse tree $t$, a complex task that will have a large impact on discontinous parsing performance, as observed by e.g. \citet{vilares-gomez-rodriguez-2020-discontinuous}, who recently dealt with reordering to extend their sequence-tagging encoding for discontinuous parsing.

In Figure~\ref{fig:trees}, we depict how a discontinuous tree (a) is converted into a continuous variant (b) by applying function $f$ to map each original position to its corresponding CCA position (c).

\subsection{Pointer Networks}
To implement function $f$ and accurately obtain the CCA positions for each token,
we rely on \textit{Pointer Networks} \cite{Vinyals15}. This neural architecture was developed to, given an input sequence, output a sequence of discrete numbers that correspond to positions from the input. Unlike regular sequence-to-sequence models that use the same dictionary of output labels for the whole training dataset, Pointer Networks employ an attention mechanism \cite{Bahdanau2014} to select positions from the input, so they can handle as many labels as the length of each sentence instead of having a fixed output dictionary size.

For our purpose, the input sequence will be 
$\mathbf{w}$ and the output sequence, the absolute CCA positions (i.e., positions $j$ in $\mathbf{w'}$). Additionally, we keep track of already-assigned CCA positions and extend the Pointer Network with the \textit{uniqueness constraint}: once a CCA position is assigned to an input token, it is no longer available for the rest of the sentence. As a consequence, the Pointer Network will just need $n$-1 steps to relocate each token of the original sentence from left to right, assigning to the last token the remaining CCA position. 

Although the overall performance of the pointer is high enough, we note that the specific accuracy on tokens affected by discontinuities is substantially lower. This was expected due to the complexity of the task and can be explained by the fact that these kind of tokens are less frequent in the training dataset and, in languages such as English, the amount of discontinuous sentences is scarce, not providing enough examples to adequately train the pointer. To increase the pointer performance, we decided to jointly train a labeller in charge of identifying those tokens. More specifically, we consider that a token is involved in a discontinuity if its original position $i$ differs from the CCA position $j$. 
This is regardless of whether the token is part of a discontinuous constituent or not, e.g., in Figure~\ref{fig:trees} it includes both the tokens in blue (that move left) and those in red (that move right). The idea behind this strategy is to prefer those models that better relocate tokens that change its absolute position in the resulting CCA.

While it can be argued that directly handling absolute CCA positions might underperform approaches that use relative positions instead (as reported by \citet{vilares-gomez-rodriguez-2020-discontinuous}), we already explored that strategy and found that the use of relative CCA positions yielded worse accuracy in a Pointer Network framework. This can be mainly explained by the fact that we cannot apply the uniqueness constraint when relative positions are used, not reducing the search space while the sentence processing advances. Moreover, in regular sequence-to-sequence approaches, the use of relative positions leads to a lower size of the output dictionary, but this benefit has no impact in Pointer Networks since the size of the dictionary will always be the sentence length.

\subsection{Neural Architecture}
Following other pointer-network-based models \cite{Ma18,L2RPointer}, we design a specific neural architecture for our problem:
\paragraph{Encoder} Each input sentence $\mathbf{w}$ is encoded, token by token, by a BiLSTM-CNN architecture \cite{Ma2016} into a sequence of \textit{encoder hidden states} $\mathbf{h}_0, \dots, \mathbf{h}_{n-1}$. To that end, each input token is initially represented as the concatenation of three different vectors obtained from character-level representations, regular pre-trained word embeddings and fixed 
contextualized word embeddings extracted from the pre-trained language model 
BERT \cite{devlin-etal-2019-bert}.

\paragraph{Decoder} An
LSTM is used to model the 
decoding process. At each time step $t$, the decoder is fed 
the encoder hidden state $\mathbf{h}_i$ of the current token $w_i$ to be relocated and generates a \textit{decoder hidden state} $\mathbf{s}_t$ that will be used for computing the probability distribution over all available CCA positions from the input (i.e., $j \in [0,n-1] \setminus A$, with $A$ being the set of already-assigned CCA positions). A biaffine scoring function \cite{DozatM17} is used for computing this probability distribution that will implement the attention mechanism:
\begin{gather*}
\mathbf{v}_{tj} = \mathbf{score}(\mathbf{s}_t, \mathbf{h}_j)= g_1(\mathbf{s}_t)^T \mathbf{W} g_2(\mathbf{h}_j)\\
+\mathbf{U}^Tg_1(\mathbf{s}_t) + \mathbf{V}^Tg_2(\mathbf{h}_j) + \mathbf{b};\\
\mathbf{a}_t = \mathbf{softmax}(\mathbf{v}_t)
\end{gather*}
\noindent where $\mathbf{W}$, $\mathbf{U}$ and $\mathbf{V}$ are the weights and $g_1(\cdot)$ and $g_2(\cdot)$ are 
multilayer perceptrons (MLP).

The attention vector $\mathbf{a}_t$ is then used as a pointer that, at time step $t$, will select the highest-scoring position $j$ as the new CCA position for the token originally located at $i$.

\begin{figure*}
\centering
\includegraphics[width=0.7\textwidth]{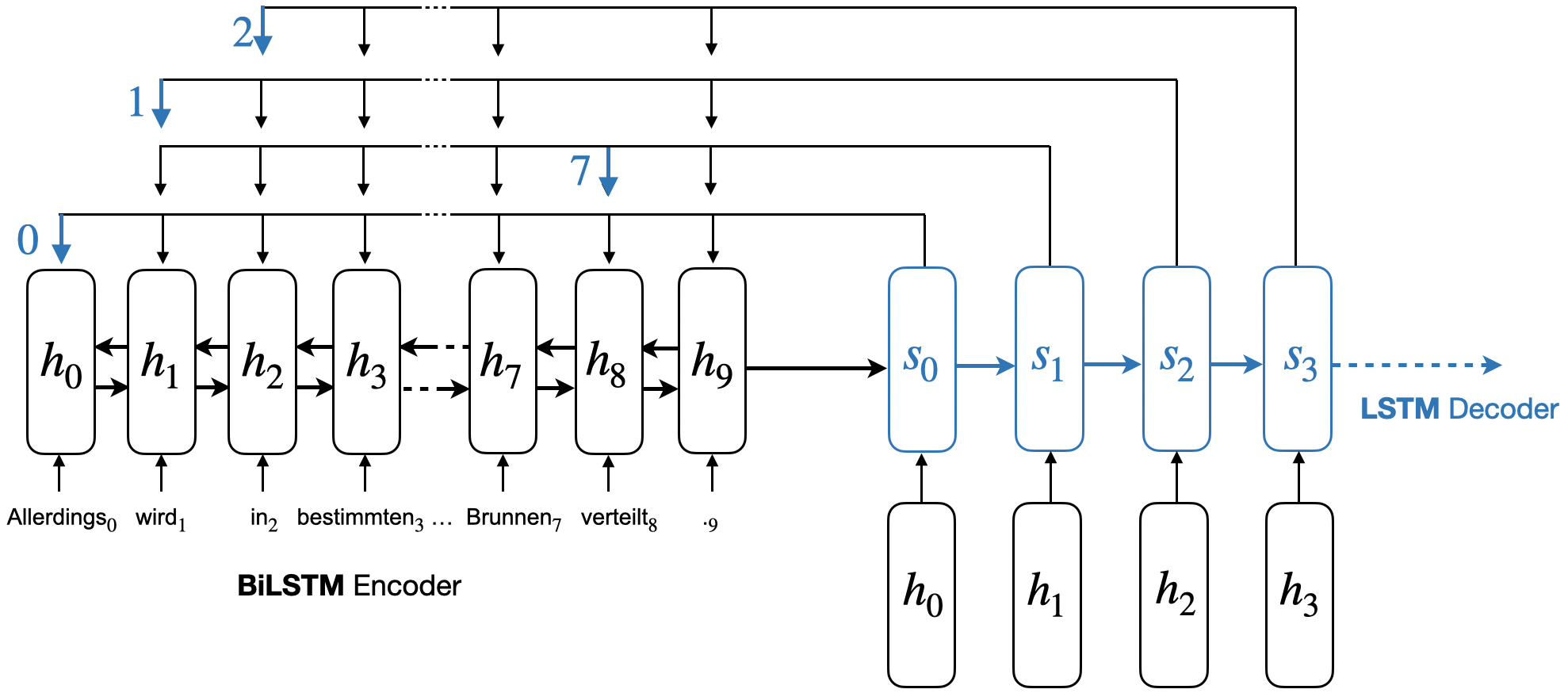}
\caption{Simplified sketch of the Pointer Network architecture and decoding steps to 
reorder the sentence in Figure~\ref{fig:trees}(a).}
\label{fig:network}
\end{figure*}

The Pointer Network is trained by minimizing the total log loss (cross entropy) to choose the correct sequence of CCA positions. Additionally, a binary biaffine classifier \cite{DozatM17} that identifies relocated tokens is jointly trained by summing the pointer and labeller losses. Since the decoding process requires $n-1$ steps to assign the CCA position to each token and at each step the attention vector $\mathbf{a}_t$ is computed over the whole input, the proposed neural model can process a sentence in 
$O(n^2)$
time complexity. Figure~\ref{fig:network} depicts
the neural architecture and the decoding procedure for reordering the sentence in Figure~\ref{fig:trees}(a).

\begin{table}
\small
\centering
\begin{tabular}{@{\hskip 0pt}l@{\hskip 2pt}c@{\hskip 4pt}c@{\hskip 6pt}c@{\hskip 6pt}c@{\hskip 6pt}c@{\hskip 4pt}c@{\hskip 4pt}c@{\hskip 0pt}}
 & \textbf{label.} & \textbf{UAS} & \textbf{Rec.} & \textbf{Prec.} & \textbf{F1} & \textbf{\%gold} & \textbf{\%pred} \\
\hline
TIGER & no & 94.16 & 76.11 & \textbf{76.20} & 76.15 & 17.65 & 17.63 \\
      & yes & \textbf{94.19} & \textbf{77.66} & 76.14 & \textbf{76.89} & 17.65 & 18.00 \\
\hdashline[1pt/1pt]
NEGRA & no & 94.56 & 79.44 & 80.20 & 79.82 & 18.73 & 18.55 \\
      & yes & \textbf{94.82} & \textbf{81.58} & \textbf{80.21} & \textbf{80.89} & 18.73 & 19.05 \\
\hdashline[1pt/1pt]
DPTB & no & 97.69 & 78.63 & 77.56 & 78.09 & 7.25 & 7.35 \\
      & yes & \textbf{97.88} & \textbf{82.20} & \textbf{78.40} & \textbf{80.26} & 7.25 & 7.61 \\
\hline
\end{tabular}
\centering
\setlength{\abovecaptionskip}{4pt}
\caption{Pointer performance with and without the labeller on dev splits. \textbf{label.} = using labeller or not. \textbf{UAS} = Unlabelled Attachment Score.  \textbf{Rec.}, \textbf{Prec.} and \textbf{F1} = Recall, precision and F-score on relocated tokens. \textbf{\%gold} = \% of relocated tokens on the gold set. \textbf{\%pred} = \% of relocated tokens predicted by the pointer.} 
\label{tab:dev}
\end{table}

\section{Experiments}
\label{sec:experiments}
\subsection{Setup}
\paragraph{Data} We test our approach 
on two 
German discontinuous treebanks, NEGRA \cite{Skut1997} and TIGER \cite{brants02}, 
and the discontinuous 
English Penn Treebank (DPTB) \cite{evang-kallmeyer-2011-plcfrs} with 
standard splits as described in Appendix~\ref{sec:splits},
discarding PoS tags in all cases. 
We apply \texttt{discodop}\footnote{ \url{https://github.com/andreasvc/disco-dop}} \cite{Cranenburgh2016} 
to transform them into continuous treebanks.  \added{This tool follows a depth-first in-order traversal that reorders words to remove crossing branches. For all treebanks, we convert
discontinuous trees in \texttt{export} format into continuous variants in \texttt{discbracket} format, using the resulting word permutation as CCAs for training the pointer and keeping track of the original word order for implementing the inverse function $f^{-1}$. Additionally, the resulting continuous treebanks in \texttt{discbracket} format are also converted by \texttt{discodop} into the commonly-used \texttt{bracket} format
for training continuous parsers.}

\paragraph{Pointer settings} Word vectors are initialized with a concatenation of pre-trained 
structured-skipgram 
embeddings \cite{Ling2015} and 
fixed weights extracted from one or several layers of the \textsc{Base} and \textsc{Large} sizes of
the pre-trained language model 
BERT \cite{devlin-etal-2019-bert}. In particular, we follow \cite{multipointer} and extract weights from the second-to-last layer for the \textsc{Base} models and, for the \textsc{Large} models, we use a combination of four layers from 17 to 20. We do not try other variations that might probably work better for our specific task. While regular word embeddings are fine-tuned during training, BERT-based embeddings are kept fixed following a less resource-consuming strategy. See Appendix~\ref{sec:appendix} for further details.

\paragraph{Parsers} For parsing the CCAs generated by the pointer, we employ two off-the-shelf continuous constituent parsers that excel in continuous benchmarks:
the  
chart-based parser by \citet{kitaev-etal-2019-multilingual} and the
transition-based model by \citet{attachjuxtapose}. In both cases, we adopt the basic configuration 
(described in their respective papers) 
and just vary the encoder initialization with 
BERT$_\textsc{Base}$ and BERT$_\textsc{Large}$ \cite{devlin-etal-2019-bert}, as well as XLNet \cite{XLNet}.

\paragraph{Metrics} Following standard practice, we ignore punctuation and root symbols for evaluating discontinuous parsing and use \texttt{discodop} for reporting F-score and discontinuous F-score (DF1).\footnote{F-score measured only on discontinuous constituents.} For jointly evaluating the pointer and labeller performance, we rely on the Labelled Attachment Score\footnote{A standard metric used for dependency parsing that, in our case, measures which tokens have the correct CCA position and also were correctly identified as relocated token.} (LAS) and choose the model with the highest score on the development set. For reporting speeds, we use sentences per second (sent/s).

\begin{table*}[tbp]
\small
\centering
\begin{tabular}{@{\hskip 0pt}l@{\hskip 3pt}c@{\hskip 6pt}c@{\hskip 6pt}cc@{\hskip 6pt}c@{\hskip 6pt}cc@{\hskip 6pt}c@{\hskip 6pt}c@{\hskip 0pt}}
& \multicolumn{3}{c}{\textbf{TIGER}}
& \multicolumn{3}{c}{\textbf{NEGRA}}
& \multicolumn{3}{c}{\textbf{DPTB}}
\\
\textbf{Parser} {\tiny (no tags or predicted PoS tags)} & \textbf{F1} & \textbf{DF1} & \textbf{sent/s} & \textbf{F1} & \textbf{DF1} & \textbf{sent/s} & \textbf{F1} & \textbf{DF1} & \textbf{sent/s} \\
\hline
\citet{Versley2016} & \added{79.5} & - & - & - & - & - & - & - & - \\
\citet{coavoux2019b} & 82.5 & 55.9 & 64 & 83.2 & 56.3 & - & 90.9 & 67.3 & 38 \\
\citet{coavoux2019a} & 82.7 & 55.9 & 126 & 83.2 & 54.6 & - & 91.0 & 71.3 & 80 \\
\citet{stanojevic-steedman-2020-span} & 83.4 & 53.5 & - & 83.6 & 50.7 & - & 90.5 & 67.1 & - \\
\citet{Corro2020SpanbasedDC} & 85.2 & 51.2 & 474 & 86.3 & 56.1 & 478 & 92.9 & 64.9 & 355 \\
\citet{Corro2020SpanbasedDC} + BERT$_\textsc{X}$ & 90.0 & 62.1 & - & 91.6 & 66.1 & - & 94.8 & 68.9 & - \\
\citet{morbitz2020supertaggingbasedNAACL} & 82.5 & 55.9 & 101 & 82.7 & 49.0 & 136 & 90.1 & 72.9 & 95    \\
\citet{morbitz2020supertaggingbasedNAACL} + BERT$_\textsc{Base}$ & 88.3 & 69.0 & 60 & 90.9 & 72.6 & 68 & 93.3 & \textbf{80.5} & 57 \\
\citet{vilares-gomez-rodriguez-2020-discontinuous} & 77.5 & 39.5 & \textbf{568} & 75.6 & 34.6 & \textbf{715} & 88.8 & 45.8 & \textbf{611} \\
\citet{vilares-gomez-rodriguez-2020-discontinuous} + BERT$_\textsc{Base}$ & 84.6 & 51.1 & 80 & 83.9 & 45.6 & 80 & 91.9 & 50.8 & 80 \\
\citet{vilares-gomez-rodriguez-2020-discontinuous} + BERT$_\textsc{Large}$ & - & - & - & - & - & - & 92.8 & 53.9 & 34 \\
\citet{DiscoPointer} & 85.7 & 60.4 & - & 85.7 & 58.6 & - & - & - & -  \\
\citet{multipointer}$^*$ & 86.6 & 62.6 & - & 86.8 & 69.5 & - & - & - & -  \\
\citet{multipointer}+BERT$_\textsc{Base}^*$ & 89.8 & \textbf{71.0} & - & 91.0 & \textbf{76.6} & - & - & - & -  \\
\hdashline[1pt/1pt]
Pointer + \citet{kitaev-etal-2019-multilingual} + BERT$_\textsc{Base}$ & 88.5 & 63.0 & 238 & 90.0 & 65.9 & 275 & 94.0 & 68.9 & 231 \\
Pointer + \citet{kitaev-etal-2019-multilingual} + BERT$_\textsc{Large}$ & \textbf{90.5} & 68.1 & 207 & \textbf{92.0} & 67.9 & 216 & 94.7 & 72.9 & 193 \\
Pointer + \citet{kitaev-etal-2019-multilingual} +  XLNet & - & - & - & - & - & - & 95.1 & 74.1 & 179 \\
Pointer + \citet{attachjuxtapose} + BERT$_\textsc{Base}$ & 88.5 & 62.7 & 157 & 90.4 & 66.5  & 188 & 94.1 & 67.2 & 152  \\
Pointer + \citet{attachjuxtapose} + BERT$_\textsc{Large}$ & \textbf{90.5} & 68.8 & 129 & 91.7 & 67.9 & 158 & 94.8 & 71.3 & 135  \\
Pointer + \citet{attachjuxtapose} + XLNet & - & - & - & - & - & - & \textbf{95.5} & 73.4 & 133 \\
\hline
\end{tabular}
\centering
\setlength{\abovecaptionskip}{4pt}
\caption{
Comparison 
of our approach 
against 
discontinuous 
constituent 
parsers on the test split.  
BERT$_\textsc{X}$ denotes that the size 
was not specified in the 
original 
paper. In those cases where the parser uses XLNet, the pointer is initialized with BERT$_\textsc{Large}$. $^*$ uses extra dependency information. \added{All reported speeds were  measured on GPU.}
}
\label{tab:disc}
\end{table*}

\subsection{Results}
Table~\ref{tab:dev} highlights how the labeller enhances the pointer's performance. While the overall UAS (or LAS when the labeller is used, since its accuracy is 100\% in all cases) is not affected substantially, it can be seen that the percentage of relocated tokens predicted by the pointer is higher (increased recall without harming precision), also leading to an improvement in F-score.

In Table~\ref{tab:disc}, we show how our novel neural architecture (combined with two continuous parsers) achieves competitive accuracies in all datasets, outperforming all existing parsers when the largest pre-trained models
are employed. It is important also to remark that F-scores on discontinuities produced by our setup (and where the pointer has an important role) are on par with purely discontinuous parsers.\footnote{\added{We believe that the remarkable performance obtained on discontinuities by \citet{multipointer} probably owes to the leverage of additional non-projective dependency information.}} Regarding efficiency, the proposed Pointer Network provides high speeds even with BERT$_\textsc{Large}$:
on the test splits,  553.7 sent/s on TIGER, 613.5 sent/s on NEGRA and 694.3 sent/s on DPTB. As a result, continuous parsers' efficiency is not penalized, and the pointer+parser combinations are faster than all existing approaches that use pre-trained language models (including the fastest parser to date by \citet{vilares-gomez-rodriguez-2020-discontinuous}, which is also outperformed by a wide margin in terms of accuracy). Finally, as also observed on continuous treebanks, no meaningful differences can be seen between both continuous parsers' performance.

\section{Conclusions and Future work}
We show that, by accurately 
removing crossing branches from discontinuous trees,
continuous parsers can perform discontinuous 
parsing more efficiently, achieving accuracies on par with more expensive 
discontinuous approaches. 
In addition, the proposed Pointer Network can be easily combined with any off-the-self continuous parser and, while barely affecting its efficiency, it can 
extend its coverage to fully model discontinuous phenomena. 

\added{We will investigate alternatives to the in-order reordering (e.g., pre- and post-order traversal or language-specific rules to generate more continuous-friendly structures). While we think that using a different CCA would have no substantial impact on Pointer Network reordering,
it might affect continuous parsing performance (as it may be easier for the parser to process reordered constituent trees with a syntax closer to original continuous structures, and factors like the degree of left vs. right branching may also have an influence).}

\section*{Acknowledgments}
We acknowledge the European Research Council (ERC), which has funded this research under the European Union’s Horizon 2020 research and innovation programme (FASTPARSE, grant agreement No 714150), ERDF/MICINN-AEI (ANSWER-ASAP, TIN2017-85160-C2-1-R), Xunta de Galicia (ED431C 2020/11), and Centro de Investigaci\'on de Galicia ``CITIC'', funded by Xunta de Galicia and the European Union (ERDF - Galicia 2014-2020 Program), by grant ED431G 2019/01.

\bibliography{anthology,main,bibliography}

\begin{thebibliography}{39}
\expandafter\ifx\csname natexlab\endcsname\relax\def\natexlab#1{#1}\fi

\bibitem[{Bahdanau et~al.(2014)Bahdanau, Cho, and Bengio}]{Bahdanau2014}
Dzmitry Bahdanau, Kyunghyun Cho, and Yoshua Bengio. 2014.
\newblock Neural machine translation by jointly learning to align and
  translate.
\newblock \emph{CoRR}, abs/1409.0473.

\bibitem[{Boyd(2007)}]{boyd-2007-discontinuity}
Adriane Boyd. 2007.
\newblock \href {https://www.aclweb.org/anthology/W07-1506} {Discontinuity
  revisited: An improved conversion to context-free representations}.
\newblock In \emph{Proceedings of the Linguistic Annotation Workshop}, pages
  41--44, Prague, Czech Republic. Association for Computational Linguistics.

\bibitem[{Brants et~al.(2002)Brants, Dipper, Hansen, Lezius, and
  Smith}]{brants02}
Sabine Brants, Stefanie Dipper, Silvia Hansen, Wolfgang Lezius, and George
  Smith. 2002.
\newblock {TIGER} treebank.
\newblock In \emph{Proceedings of the 1st Workshop on Treebanks and Linguistic
  Theories (TLT)}, pages 24--42.

\bibitem[{Coavoux and Cohen(2019)}]{coavoux2019b}
Maximin Coavoux and Shay~B. Cohen. 2019.
\newblock Discontinuous constituency parsing with a stack-free transition
  system and a dynamic oracle.
\newblock In \emph{Proceedings of the 2019 Conference of the North {A}merican
  Chapter of the Association for Computational Linguistics}, pages 204--217,
  Minneapolis, Minnesota. Association for Computational Linguistics.

\bibitem[{Coavoux and Crabb{\'e}(2017)}]{coavoux2017}
Maximin Coavoux and Beno{\^\i}t Crabb{\'e}. 2017.
\newblock Incremental discontinuous phrase structure parsing with the {GAP}
  transition.
\newblock In \emph{Proceedings of the 15th Conference of the {E}uropean Chapter
  of the Association for Computational Linguistics}, pages 1259--1270,
  Valencia, Spain. Association for Computational Linguistics.

\bibitem[{Coavoux et~al.(2019)Coavoux, Crabb{\'e}, and Cohen}]{coavoux2019a}
Maximin Coavoux, Beno{\^\i}t Crabb{\'e}, and Shay~B. Cohen. 2019.
\newblock Unlexicalized transition-based discontinuous constituency parsing.
\newblock \emph{Transactions of the Association for Computational Linguistics},
  7:73--89.

\bibitem[{Corro(2020)}]{Corro2020SpanbasedDC}
Caio Corro. 2020.
\newblock \href {https://doi.org/10.18653/v1/2020.emnlp-main.219} {Span-based
  discontinuous constituency parsing: a family of exact chart-based algorithms
  with time complexities from {O}(n{\^{}}6) down to {O}(n{\^{}}3)}.
\newblock In \emph{Proceedings of the 2020 Conference on Empirical Methods in
  Natural Language Processing (EMNLP)}, pages 2753--2764, Online. Association
  for Computational Linguistics.

\bibitem[{Devlin et~al.(2019)Devlin, Chang, Lee, and
  Toutanova}]{devlin-etal-2019-bert}
Jacob Devlin, Ming-Wei Chang, Kenton Lee, and Kristina Toutanova. 2019.
\newblock \href {https://doi.org/10.18653/v1/N19-1423} {{BERT}: Pre-training of
  deep bidirectional transformers for language understanding}.
\newblock In \emph{Proceedings of the 2019 Conference of the North {A}merican
  Chapter of the Association for Computational Linguistics: Human Language
  Technologies, Volume 1 (Long and Short Papers)}, pages 4171--4186,
  Minneapolis, Minnesota. Association for Computational Linguistics.

\bibitem[{Dozat and Manning(2017)}]{DozatM17}
Timothy Dozat and Christopher~D. Manning. 2017.
\newblock Deep biaffine attention for neural dependency parsing.
\newblock In \emph{{ICLR}}. OpenReview.net.

\bibitem[{Dubey and Keller(2003)}]{dubey2003}
Amit Dubey and Frank Keller. 2003.
\newblock Probabilistic parsing for {G}erman using sister-head dependencies.
\newblock In \emph{Proceedings of the 41st Annual Meeting of the Association
  for Computational Linguistics}, pages 96--103, Sapporo, Japan.

\bibitem[{Evang and Kallmeyer(2011)}]{evang-kallmeyer-2011-plcfrs}
Kilian Evang and Laura Kallmeyer. 2011.
\newblock \href {https://www.aclweb.org/anthology/W11-2913} {{PLCFRS} parsing
  of {E}nglish discontinuous constituents}.
\newblock In \emph{Proceedings of the 12th International Conference on Parsing
  Technologies}, pages 104--116, Dublin, Ireland. Association for Computational
  Linguistics.

\bibitem[{Fern{\'a}ndez-Gonz{\'a}lez and
  G{\'o}mez-Rodr{\'\i}guez(2019)}]{L2RPointer}
Daniel Fern{\'a}ndez-Gonz{\'a}lez and Carlos G{\'o}mez-Rodr{\'\i}guez. 2019.
\newblock \href {https://doi.org/10.18653/v1/N19-1076} {Left-to-right
  dependency parsing with pointer networks}.
\newblock In \emph{Proceedings of the 2019 Conference of the North {A}merican
  Chapter of the Association for Computational Linguistics: Human Language
  Technologies, Volume 1 (Long and Short Papers)}, pages 710--716, Minneapolis,
  Minnesota. Association for Computational Linguistics.

\bibitem[{Fern{\'{a}}ndez{-}Gonz{\'{a}}lez and
  G{\'{o}}mez{-}Rodr{\'{i}}guez(2020{\natexlab{a}})}]{DiscoPointer}
Daniel Fern{\'{a}}ndez{-}Gonz{\'{a}}lez and Carlos
  G{\'{o}}mez{-}Rodr{\'{i}}guez. 2020{\natexlab{a}}.
\newblock \href {https://doi.org/https://doi.org/10.1609/aaai.v34i05.6275}
  {Discontinuous constituent parsing with pointer networks}.
\newblock In \emph{Proceedings of the Thirty-Fourth {AAAI} Conference on
  Artificial Intelligence, {AAAI} 2020, New York, NY, USA, February 7-12,
  2020}, pages 7724--7731. {AAAI} Press.

\bibitem[{Fern{\'{a}}ndez{-}Gonz{\'{a}}lez and
  G{\'{o}}mez{-}Rodr{\'{i}}guez(2020{\natexlab{b}})}]{multipointer}
Daniel Fern{\'{a}}ndez{-}Gonz{\'{a}}lez and Carlos
  G{\'{o}}mez{-}Rodr{\'{i}}guez. 2020{\natexlab{b}}.
\newblock \href {http://arxiv.org/abs/2009.09730} {Multitask pointer network
  for multi-representational parsing}.

\bibitem[{Fern{\'a}ndez-Gonz{\'a}lez and
  Martins(2015)}]{fernandez-gonzalez-martins-2015-parsing}
Daniel Fern{\'a}ndez-Gonz{\'a}lez and Andr{\'e} F.~T. Martins. 2015.
\newblock \href {https://doi.org/10.3115/v1/P15-1147} {Parsing as reduction}.
\newblock In \emph{Proceedings of the 53rd Annual Meeting of the Association
  for Computational Linguistics and the 7th International Joint Conference on
  Natural Language Processing (Volume 1: Long Papers)}, pages 1523--1533,
  Beijing, China. Association for Computational Linguistics.

\bibitem[{Gebhardt(2020)}]{gebhardt-2020-advances}
Kilian Gebhardt. 2020.
\newblock \href {https://doi.org/10.18653/v1/2020.iwpt-1.9} {Advances in using
  grammars with latent annotations for discontinuous parsing}.
\newblock In \emph{Proceedings of the 16th International Conference on Parsing
  Technologies and the IWPT 2020 Shared Task on Parsing into Enhanced Universal
  Dependencies}, pages 91--97, Online. Association for Computational
  Linguistics.

\bibitem[{Kingma and Ba(2014)}]{Adam}
Diederik~P. Kingma and Jimmy Ba. 2014.
\newblock Adam: A method for stochastic optimization.
\newblock Published as a conference paper at the 3rd International Conference
  for Learning Representations, San Diego, 2015.

\bibitem[{Kitaev et~al.(2019)Kitaev, Cao, and
  Klein}]{kitaev-etal-2019-multilingual}
Nikita Kitaev, Steven Cao, and Dan Klein. 2019.
\newblock \href {https://doi.org/10.18653/v1/P19-1340} {Multilingual
  constituency parsing with self-attention and pre-training}.
\newblock In \emph{Proceedings of the 57th Annual Meeting of the Association
  for Computational Linguistics}, pages 3499--3505, Florence, Italy.
  Association for Computational Linguistics.

\bibitem[{Ling et~al.(2015)Ling, Dyer, Black, and Trancoso}]{Ling2015}
Wang Ling, Chris Dyer, Alan~W. Black, and Isabel Trancoso. 2015.
\newblock Two/too simple adaptations of {W}ord2{V}ec for syntax problems.
\newblock In \emph{Proceedings of the 2015 Conference of the North {A}merican
  Chapter of the Association for Computational Linguistics}, pages 1299--1304,
  Denver, Colorado.

\bibitem[{Ma and Hovy(2016)}]{Ma2016}
Xuezhe Ma and Eduard Hovy. 2016.
\newblock End-to-end sequence labeling via bi-directional lstm-cnns-crf.
\newblock In \emph{Proceedings of the 54th Annual Meeting of the Association
  for Computational Linguistics}, pages 1064--1074. Association for
  Computational Linguistics.

\bibitem[{Ma et~al.(2018)Ma, Hu, Liu, Peng, Neubig, and Hovy}]{Ma18}
Xuezhe Ma, Zecong Hu, Jingzhou Liu, Nanyun Peng, Graham Neubig, and Eduard~H.
  Hovy. 2018.
\newblock Stack-pointer networks for dependency parsing.
\newblock In \emph{Proceedings of the 56th Annual Meeting of the Association
  for Computational Linguistics, Melbourne, Australia, July 15-20, 2018}, pages
  1403--1414.

\bibitem[{Maier(2015)}]{maier-2015-discontinuous}
Wolfgang Maier. 2015.
\newblock \href {https://doi.org/10.3115/v1/P15-1116} {Discontinuous
  incremental shift-reduce parsing}.
\newblock In \emph{Proceedings of the 53rd Annual Meeting of the Association
  for Computational Linguistics and the 7th International Joint Conference on
  Natural Language Processing (Volume 1: Long Papers)}, pages 1202--1212,
  Beijing, China. Association for Computational Linguistics.

\bibitem[{Maier and Lichte(2016)}]{maier2016}
Wolfgang Maier and Timm Lichte. 2016.
\newblock Discontinuous parsing with continuous trees.
\newblock In \emph{Proceedings of the Workshop on Discontinuous Structures in
  Natural Language Processing}, pages 47--57, San Diego, California.
  Association for Computational Linguistics.

\bibitem[{Marcus et~al.(1993)Marcus, Santorini, and Marcinkiewicz}]{marcus93}
Mitchell~P. Marcus, Beatrice Santorini, and Mary~Ann Marcinkiewicz. 1993.
\newblock Building a large annotated corpus of {E}nglish: The {P}enn
  {T}reebank.
\newblock \emph{Computational Linguistics}, 19:313--330.

\bibitem[{Nivre and Nilsson(2005)}]{nivre-nilsson-2005-pseudo}
Joakim Nivre and Jens Nilsson. 2005.
\newblock \href {https://doi.org/10.3115/1219840.1219853} {Pseudo-projective
  dependency parsing}.
\newblock In \emph{Proceedings of the 43rd Annual Meeting of the Association
  for Computational Linguistics ({ACL}{'}05)}, pages 99--106, Ann Arbor,
  Michigan. Association for Computational Linguistics.

\bibitem[{Ruprecht and M{\"o}rbitz(2021)}]{morbitz2020supertaggingbasedNAACL}
Thomas Ruprecht and Richard M{\"o}rbitz. 2021.
\newblock \href {https://doi.org/10.18653/v1/2021.naacl-main.232}
  {Supertagging-based parsing with linear context-free rewriting systems}.
\newblock In \emph{Proceedings of the 2021 Conference of the North American
  Chapter of the Association for Computational Linguistics: Human Language
  Technologies}, pages 2923--2935, Online. Association for Computational
  Linguistics.

\bibitem[{Seddah et~al.(2013)Seddah, Tsarfaty, K{\"u}bler, Candito, Choi,
  Farkas, Foster, Goenaga, Gojenola~Galletebeitia, Goldberg, Green, Habash,
  Kuhlmann, Maier, Nivre, Przepi{\'o}rkowski, Roth, Seeker, Versley, Vincze,
  Woli{\'n}ski, Wr{\'o}blewska, and Villemonte de~la
  Clergerie}]{seddah-etal-2013-overview}
Djam{\'e} Seddah, Reut Tsarfaty, Sandra K{\"u}bler, Marie Candito, Jinho~D.
  Choi, Rich{\'a}rd Farkas, Jennifer Foster, Iakes Goenaga, Koldo
  Gojenola~Galletebeitia, Yoav Goldberg, Spence Green, Nizar Habash, Marco
  Kuhlmann, Wolfgang Maier, Joakim Nivre, Adam Przepi{\'o}rkowski, Ryan Roth,
  Wolfgang Seeker, Yannick Versley, Veronika Vincze, Marcin Woli{\'n}ski, Alina
  Wr{\'o}blewska, and Eric Villemonte de~la Clergerie. 2013.
\newblock \href {https://www.aclweb.org/anthology/W13-4917} {Overview of the
  {SPMRL} 2013 shared task: A cross-framework evaluation of parsing
  morphologically rich languages}.
\newblock In \emph{Proceedings of the Fourth Workshop on Statistical Parsing of
  Morphologically-Rich Languages}, pages 146--182, Seattle, Washington, USA.
  Association for Computational Linguistics.

\bibitem[{Skut et~al.(1997)Skut, Krenn, Brants, and Uszkoreit}]{Skut1997}
Wojciech Skut, Brigitte Krenn, Thorsten Brants, and Hans Uszkoreit. 1997.
\newblock An annotation scheme for free word order languages.
\newblock In \emph{Proceedings of the Fifth Conference on Applied Natural
  Language Processing}, ANLC '97, pages 88--95, Stroudsburg, PA, USA.
  Association for Computational Linguistics.

\bibitem[{Stanojevi{\'c} and Alhama(2017)}]{stanojevic-alhama-2017-neural}
Milo{\v{s}} Stanojevi{\'c} and Raquel~G. Alhama. 2017.
\newblock \href {https://doi.org/10.18653/v1/D17-1174} {Neural discontinuous
  constituency parsing}.
\newblock In \emph{Proceedings of the 2017 Conference on Empirical Methods in
  Natural Language Processing}, pages 1666--1676, Copenhagen, Denmark.
  Association for Computational Linguistics.

\bibitem[{Stanojevi{\'c} and Steedman(2020)}]{stanojevic-steedman-2020-span}
Milo{\v{s}} Stanojevi{\'c} and Mark Steedman. 2020.
\newblock \href {https://doi.org/10.18653/v1/2020.iwpt-1.12} {Span-based
  {LCFRS}-2 parsing}.
\newblock In \emph{Proceedings of the 16th International Conference on Parsing
  Technologies and the IWPT 2020 Shared Task on Parsing into Enhanced Universal
  Dependencies}, pages 111--121, Online. Association for Computational
  Linguistics.

\bibitem[{Stern et~al.(2017)Stern, Andreas, and
  Klein}]{stern-etal-2017-minimal}
Mitchell Stern, Jacob Andreas, and Dan Klein. 2017.
\newblock \href {https://doi.org/10.18653/v1/P17-1076} {A minimal span-based
  neural constituency parser}.
\newblock In \emph{Proceedings of the 55th Annual Meeting of the Association
  for Computational Linguistics (Volume 1: Long Papers)}, pages 818--827,
  Vancouver, Canada. Association for Computational Linguistics.

\bibitem[{van Cranenburgh et~al.(2016)van Cranenburgh, Scha, and
  Bod}]{Cranenburgh2016}
Andreas van Cranenburgh, Remko Scha, and Rens Bod. 2016.
\newblock Data-oriented parsing with discontinuous constituents and function
  tags.
\newblock \emph{J. Language Modelling}, 4:57--111.

\bibitem[{Versley(2014)}]{versley2014}
Yannick Versley. 2014.
\newblock Experiments with easy-first nonprojective constituent parsing.
\newblock In \emph{Proceedings of the First Joint Workshop on Statistical
  Parsing of Morphologically Rich Languages and Syntactic Analysis of
  Non-Canonical Languages}, pages 39--53, Dublin, Ireland. Dublin City
  University.

\bibitem[{Versley(2016)}]{Versley2016}
Yannick Versley. 2016.
\newblock Discontinuity (re){\mbox{$^2$}}-visited: A minimalist approach to
  pseudoprojective constituent parsing.
\newblock In \emph{Proceedings of the Workshop on Discontinuous Structures in
  Natural Language Processing}, pages 58--69, San Diego, California.

\bibitem[{Vijay-Shanker et~al.(1987)Vijay-Shanker, Weir, and
  Joshi}]{VijWeiJos87}
K.~Vijay-Shanker, David~J. Weir, and Aravind~K. Joshi. 1987.
\newblock Characterizing structural descriptions produced by various
  grammatical formalisms.
\newblock In \emph{Proceedings of the 25th Annual Meeting of the Association
  for Computational Linguistics (ACL'87)}, pages 104--111, Morristown, NJ, USA.
  Association for Computational Linguistics.

\bibitem[{Vilares and
  G{\'o}mez-Rodr{\'\i}guez(2020)}]{vilares-gomez-rodriguez-2020-discontinuous}
David Vilares and Carlos G{\'o}mez-Rodr{\'\i}guez. 2020.
\newblock \href {https://doi.org/10.18653/v1/2020.emnlp-main.221}
  {Discontinuous constituent parsing as sequence labeling}.
\newblock In \emph{Proceedings of the 2020 Conference on Empirical Methods in
  Natural Language Processing (EMNLP)}, pages 2771--2785, Online. Association
  for Computational Linguistics.

\bibitem[{Vinyals et~al.(2015)Vinyals, Fortunato, and Jaitly}]{Vinyals15}
Oriol Vinyals, Meire Fortunato, and Navdeep Jaitly. 2015.
\newblock Pointer networks.
\newblock In C.~Cortes, N.~D. Lawrence, D.~D. Lee, M.~Sugiyama, and R.~Garnett,
  editors, \emph{Advances in Neural Information Processing Systems 28}, pages
  2692--2700. Curran Associates, Inc.

\bibitem[{Yang and Deng(2020)}]{attachjuxtapose}
Kaiyu Yang and Jia Deng. 2020.
\newblock Strongly incremental constituency parsing with graph neural networks.
\newblock In \emph{Neural Information Processing Systems}.

\bibitem[{Yang et~al.(2019)Yang, Dai, Yang, Carbonell, Salakhutdinov, and
  Le}]{XLNet}
Zhilin Yang, Zihang Dai, Yiming Yang, Jaime Carbonell, Russ~R Salakhutdinov,
  and Quoc~V Le. 2019.
\newblock \href
  {https://proceedings.neurips.cc/paper/2019/file/dc6a7e655d7e5840e66733e9ee67cc69-Paper.pdf}
  {Xlnet: Generalized autoregressive pretraining for language understanding}.
\newblock In \emph{Advances in Neural Information Processing Systems},
  volume~32. Curran Associates, Inc.

\end{thebibliography}
\bibliographystyle{acl_natbib}

\clearpage
\appendix

\section{Appendices}
\label{sec:apex}

\subsection{Treebank splits}
\label{sec:splits}
Standard splits for discontinuous German NEGRA \cite{Skut1997} and TIGER \cite{brants02} treebanks are defined by \cite{dubey2003} and \cite{seddah-etal-2013-overview}, respectively. For the discontinuous 
version of the 
English Penn Treebank (DPTB) \cite{evang-kallmeyer-2011-plcfrs},
commonly-used 
splits are defined as follows: Sections 2 to 21 for training, 22 for development and 23 for testing. In Table~\ref{tab:treebanks}, we report the number of samples per treebank split.

\begin{table}[h]
\centering
\begin{tabular}{@{\hskip 0pt}lccc@{\hskip 0pt}}
\hline
\textbf{Treebank} & \textbf{Training} & \textbf{Dev} & \textbf{Test} \\
\hline
TIGER & 40,472 & 5,000 & 5,000 \\
NEGRA & 18,602 & 1,000 & 1,000 \\
DPTB & 39,832 & 1,700 & 2,416 \\
\hline
\end{tabular}
\setlength{\abovecaptionskip}{4pt}
\caption{Treebank statistics.}
\label{tab:treebanks}
\end{table}

\subsection{Pointer Settings}
\label{sec:appendix}

\paragraph{Pre-trained word embeddings} The specific pre-trained BERT models used in this work were: for English, {\tt bert-base-cased} and {\tt bert-large-cased}; and, for German, {\tt bert-base-german-cased} and {\tt deepset/gbert-large}. The same pre-trained models were also used for both continuous parsers, including also {\tt xlnet-large-cased} in those experiments where XLNet is employed for initializing the parser's encoder.

\paragraph{Hyper-parameters} We use the Adam optimizer \cite{Adam} and the same hyper-parameter selection as the dependency parser by \cite{Ma18}. No further adaptation to our specific task was performed. These are detailed in Table~\ref{tab:hyper}. 

\paragraph{Hardware} Our approach (combined with the two specified continuous parsers) was fully tested on an Intel(R) Core(TM) i9-10920X CPU @ 3.50GHz with a single GeForce RTX 3090 GPU. All speeds are measured considering the time taken by the whole process, i.e., the reordering, continuous parsing and reversal of the reordering.

\begin{table}[h]
\begin{footnotesize}
\centering
\begin{tabular}{@{\hskip 0pt}lc@{\hskip 0pt}}
\hline
\textbf{Architecture hyper-parameters} & \\
\hline
CNN window size & 3 \\
CNN number of filters & 50 \\
BiLSTM encoder layers & 3 \\
BiLSTM encoder size & 512 \\
LSTM decoder layers & 1 \\ 
LSTM decoder size & 512 \\
LSTM layers dropout & 0.33 \\
Word/Char. embedding dimension & 100\\
BERT$_\textsc{BASE}$ embedding dimension & 768\\
BERT$_\textsc{LARGE}$ embedding dimension & 1024\\
Embeddings dropout & 0.33 \\
MLP layers & 1 \\
MLP activation function & ELU \\
CCA Position MLP size & 512 \\ 
Label MLP size & 128 \\
UNK replacement probability & 0.5 \\
Beam size & 10 \\
\hline
\textbf{Adam optimizer hyper-parameters} &\\
\hline
Initial learning rate & 0.001 \\
$\beta_1$, $\beta_2$ & 0.9 \\
Batch size & 32 \\
Decay rate & 0.75 \\
Gradient clipping & 5.0 \\
\hline
\end{tabular}
\setlength{\abovecaptionskip}{4pt}
\caption{Model hyper-parameters.}
\label{tab:hyper}
\end{footnotesize}
\end{table}

\end{document}